# Π-RT: A Runtime Framework to Enable Energy-Efficient Real-Time Robotic Applications on Heterogeneous Architectures

Liu Liu[1], Shaoshan Liu[2], Zhe Zhang, Bo Yu, Jie Tang[3], and Yuan Xie

*Abstract*— Enabling full robotic workloads with diverse behaviors on mobile systems with stringent resource and energy constraints remains a challenge. In recent years, attempts have been made to deploy single-accelerator-based computing platforms (such as GPU, DSP, or FPGA) to address this challenge, but with little success. The core problem is two-fold: firstly, different robotic tasks require different accelerators, and secondly, managing multiple accelerators simultaneously is overwhelming for developers. In this paper, we propose Π-RT, the first robotic runtime framework to efficiently manage dynamic task executions on mobile systems with multiple accelerators as well as on the cloud to achieve better performance and energy savings. With Π-RT, we enable a robot to simultaneously perform autonomous navigation with 25 FPS of localization, obstacle detection with 3 FPS, route planning, large map generation, and scene understanding, traveling at a max speed of 5 miles per hour, all within an 11W computing power envelope.

## I. INTRODUCTION

Robotic applications pose high computational and energy efficiency requirements, which are challenging for contemporary mobile systems with resource and energy constraints. One common feature of robotic applications is perception, i.e. understanding the environment and updating the location at real-time. Specifically, while moving in real-time, robots need to track their own positions and to recognize obstacles along the way. The core technologies that enable the emerging robotic applications include Computer Vision (CV), Deep Learning (DL), and Simultaneous Localization and Mapping (SLAM). Each technology is challenging and demanding on computing resources and energy budget, and all integrated form an extremely complex pipeline.

Figure 1 shows a typical robotic application pipeline. The perception of the environment is based on computer vision innovations such as image processing where the workloads are computationally intensive and often have real-time requirements. In addition, deep convolutional neural networks (CNNs) have brought a revolution in image understanding such as image labeling, which is used in autonomous mobile robots [1]. SLAM systems widely used in autonomous robotic applications tackle the problem of constructing and updating a map of an unknown environment while simultaneously keeping track of the location [2].

[1]Liu Liu and Yuan Xie are with the department of Electrical and Computer Engineering University of California, Santa Barbara, CA 93106, USA {liu_liu, yuanxie}@ucsb.edu
[2]Shaoshan Liu, Zhe Zhang, and Bo Yu are with PerceptIn Inc., Santa Clara, CA 95054, USA shaoshan.liu@perceptin.io
[3]Jie Tang is with South China University of Technology, GuangZhou, China

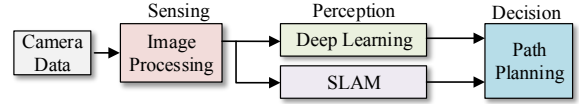

Fig. 1. A typical robotic application.

In this paper, to address the problem of enabling full robotic workloads on mobile devices, we design and implement Π-RT, a production robotic runtime framework to achieve performance improvement and energy efficiency on mobile heterogeneous computing systems as well as on the cloud. The major contributions of this work are listed as follows:

- In Section III we perform a comprehensive study of executing core robotic workloads on different heterogeneous platforms. The results demonstrate that these workloads have diverse computational behaviors and demand different accelerators.
- In Section IV, we propose Π-RT, a robotic runtime framework to manage computation offloading and resource allocation, to provide a transparent view for programmers, and to enable dynamic performance and energy consumption trade-offs.
- In Section V, we implement Π-RT on Snapdragon 820 SoC to enable an autonomous robot to simultaneously perform autonomous navigation, obstacle detection, route planning, large map generation, and scene understanding, all with an 11W of power consumption.

## II. EMERGING ROBOTIC APPLICATIONS

Emerging robotic applications act as complex systems with multiple components as indicated in Figure 1. A brief background of each component is given in this section; a real-world robotic application that integrates all these components is demonstrated in Section V.

**Embedded Computer Vision:** Computer vision (CV) algorithms usually act as the common front-end of robotic applications, which enable machines to visually sense the real-world. The representative workloads among various CV algorithms are listed in Table I. Raw image data from the sensor is first transformed by `Undistort` to compensate radial and tangential lens distortion. After `Undistort`, the `Gaussian Blur` function convolves the image using a Gaussian kernel to reduce noise. Feature detection is performed by utilizing the `GoodFeaturesToTrack` algorithm to identify the most prominent corners in images [3]. Finally, `Optical Flow` estimates the motion of objects

TABLE I
EVALUATED WORKLOADS

| Workload | Description | Ops. |
|---|---|---|
| Gaussian blur | 640x480 image 5x5 kernel | 15.4M |
| Convolution | 640x480 image 7x7 kernel | 30.1M |
| Sobel filter | 1920x1080 image | 74.7M |
| Undistort | 640x480 image | - |
| Fearture detect | 640x480 image | - |
| Optical flow | 52x52 window | - |
| CONV1 | Convolution, ReLU | 210.8M |
| CONV2 | Convolution, ReLU | 895.5M |
| CONV3 | Convolution, ReLU | 299.0M |
| CONV4 | Convolution, ReLU | 448.6M |
| CONV5 | Convolution, ReLU | 299.0M |
| FC6 | Fully-connected, dropout | 75.5M |
| FC7 | Fully-connected, dropout | 33.6M |
| FC8 | Fully-connected, softmax | 8M |

in a video stream, which is useful for object detection and tracking.

**Simultaneous Localization And Mapping:** SLAM solves the problem of constructing and updating a map of an unknown environment while simultaneously keeping track of an agent's location within it [2]. As shown in Figure 2(a), a simplified version of the general SLAM pipeline operates as follows.

The Inertial Measurement Unit (IMU) produces six data points (angular velocities in three different axis and accelerations in the three axes) at a high rate and feeds the data to the propagation stage. The Propagation Unit integrates the IMU data points and produces a new positions at real time. However, as time progresses, the position errors caused by IMU inaccuracies accumulate and cause position drifts. To correct the drift problem, a camera is used to capture frames along the path at a fixed rate, usually at 30 FPS.

The frames captured by the camera can be fed to the Feature Extraction Unit, which extracts useful corner features and generates a descriptor for each feature. The features extracted can then be fed to the local Mapping Unit to extend the map (a collection of 3D spatial points, each with a unique descriptor) as the agent keeps exploring. Meanwhile, the newly detected features are sent to the Update Unit where they are compared with the map points. If the detected features already exist in the map, the Update Unit can then derive the agent's current position from the known map points. By using this new position, the Update Unit can correct the drift introduced by the Propagation Unit. Also, the Update unit updates the map with the newly detected feature points.

**Deep Learning Algorithms:** Deep Learning is revolutionary

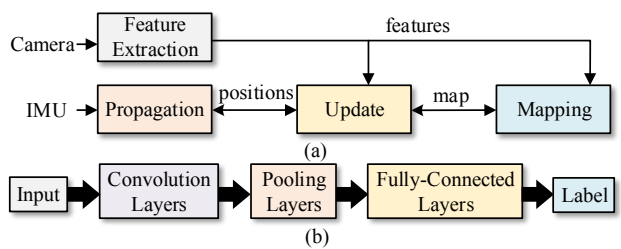

Fig. 2. (a) simplified SLAM pipeline; (b) simplified DL evaluation pipeline.

in many application scenarios including robotics. Convolutional neural networks (CNNs) is widely adopted in computer vision applications due to many practical successes in both industry and academia, which is also adopted in autonomous mobile robots [1], [4], [5], [6]. CNNs are designed and deployed to process image data as 2D arrays that contain pixel intensities in three color channels. As shown in Figure 2(b), a typical CNN is structured into layers as discussed below.

*Convolutional layer:* To extract features in local conjunction from previous layers, convolutional layers are formed by multiple feature maps each of which contains a set of trained weights referred as the kernels. After performing 2D convolutional operations on the previous layers using the kernels, the locally weighted sum is passed to an activation function such as ReLU, which then serves as the feature maps to the next layer.

*Pooling layer*: While convolutional layers extracts features from the previous layer, pooling layers merge similar features from a local patch of one feature map. The commonly used pooling method, i.e. max pooling is considered, which computes the maximum of values from a local patch in a feature map. Local patches are shifted by a stride of two or more. Therefore, the previous feature map has reduced the size and becomes invariant to small transformations.

*Fully-connected (FC) layer*: To perform final classification, FC layers contain neurons with full connections to all activations in the previous layer, and then linear transformations are applied to input feature vectors and biases.

### III. WORKLOADS CHARACTERIZATION

In this section, the behaviors of different robotic tasks on existing heterogeneous platforms are characterized. Through this study, different workloads in emerging robotic applications show a wide range of discrepancy in terms of performance and energy usage on different accelerators. Characterization on various platforms reveals that no single-accelerator architecture could accommodate all workloads.

*A. Target Accelerator Platforms*

The specifications of the accelerator-based SoC platforms are listed in Table II. **Mobile SoC** devices have been recently released to accelerate deep learning algorithms [7]. A Qualcomm Snapdragon 820 processor is investigated to show its potential for accelerating robotics applications. This mobile SoC contains a GPU core (denoted as **mGPU**) and a DSP core, which is capable of accelerating operations offloaded from CPU cores. The GPU and DSP subsystems have their private memory spaces which share the same memory controller with the CPU. Shared virtual memory (SVM) is supported by mGPU which enables addressing host data structures from mGPU by passing host pointers.

**GPU**-based SoC platforms such as NVIDIA Jetson TX1 have been widely recognized in academia and industry for the deployment of deep learning applications and embedded computer vision tasks. The architecture of TX1 features four high performance ARM Cortex-A57 cores and a 2MB L2 cache, as well as 256 Maxwell GPU cores which run at 998MHz. The GPU-based accelerators share a 256KB L2 cache and use the same memory interface to access system memory.

TABLE II
PLATFORM SPECIFICATIONS

| Platform | Snapdragon 820 | Jetson TX1 | DE1-SoC |
|---|---|---|---|
| Processor | Quad-core Kryo ARMv8 2200MHz | Quad-core ARM A57 1900MHz | Dual-core ARM A9 925MHz |
| Accelerator | Adreno 530 GPU 256 ALUs 624MHz Hexagon 680 DSP 500MHz per-thread | Maxwell GPU 256 ALUs 998MHz | Cyclone V 85K logic elements |
| Theoretical throughput | mGPU: 160 GOPS/s; DSP: 4 GOPS/s | 256 GOPS/s | |
| Memory | 3GB LPDDR4 1866MHz | 4GB LPDDR4 1600MHz | 64MB SDRAM on FPGA 1GB DDR3 SDRAM on HPS |
| Host OS | Android | Ubuntu 16.04 | Linux Console |
| Technology | 14nm FinFET LPP Samsung | 20nm SOC-TSMC | 28nm LP-TSMC |

**FPGA**-based SoC platforms are favored in embedded vision applications due to their flexibility and power efficiency [8]. Moreover, the rapid growth in computer vision algorithms makes the programmability of FPGA devices even more promising [9]. In this work, the DE1-SoC board from Terasic is used to conduct experiments, which features a dual-core ARM Cortex-A9 as the hard processor system (HPS) and a Cyclone V FPGA fabric as the programmable logic which has 85K logic elements. The FPGA device is equipped with a 64MB SDRAM as device memory. The communication between HPS and FPGA is realized by the HPS-to-FPGA AXI bridge.

*B. Experimental Setup*

The characterization study is focused on computer vision and deep learning algorithms with the evaluated workloads listed in Table I. To evaluate CV algorithms on target accelerator platforms, OpenCV CUDA image processing and feature extraction modules, vendor-supplied FPGA benchmarks, and FastCV library are used on GPU-based SoC, FPGA-based Soc, and DSP-based SoC, respectively. Based on *AlexNet* [10], the inference of CNN is evaluated on GPU-based SoC with CuDNNv5 library, and on FPGA-based SoC using hand-crafted OpenCL kernels of convolution layers and fully-connected layers.

The power measurement on TX1 is based on the INA monitors on board. The *PowerPlay Early Power Estimator* and an USB power monitor are used to generate the power consumption of FPGA and mobile SoC, respectively.

Note that CPU, DSP, and mGPU denote the host processor and accelerators on the mobile SoC, i.e. Snapdragon 820, whereas GPU and FPGA represent the accelerators on TX1 and DE1-SoC.

*C. Overall Performance and Energy*

**Observation One: Data-flow execution pattern exists.** As shown in Figure 1, robotic applications often demonstrate data-flow execution patterns such that the execution of one stage (*e.g* feature extraction) depends on the output of the previous stage (*e.g.* image capture). Resource leak occurs when the input is ready but the execution is delayed. For instance, if feature extraction does not start when images are ready, then the runtime has to either use more buffers

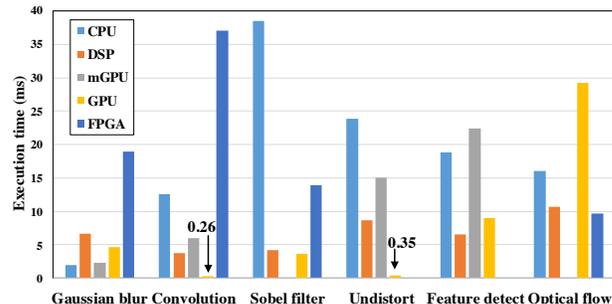
Fig. 3. Comparing the execution time per frame of CV workloads. GPU has higher performance on filter-like workloads, while DSP and FPGA shows better execution on `Feature Detect` and `Optical Flow`, respectively.

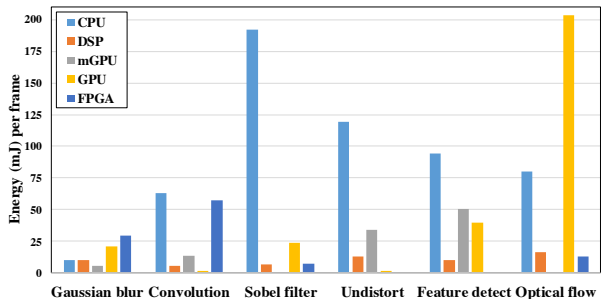
Fig. 4. Comparing the energy consumption (mJ/frame) of CV workloads. Accelerators with higher performance do not necessarily result in better energy efficiency. For example, `Sobel Filter` consumes less energy even though it has longer execution time than (m)GPU.

or to discard the images, which leads to the degradation of stability.

**Observation Two: No single accelerator wins all.** As shown in Figure 3, workloads such as `Convolution` and `Sobel Filter` with regular parallel patterns are best suited for GPU execution over other computing platforms. From the perspective of parallel computing model, such workloads are favorable in Single Instruction Multiple Thread (SIMT) execution which fits better on GPU. Similar observation can be made in DL workloads where GPU shows better performance across all layers than FPGA as shown in Figure 5. However, `Feature Detect`, with more control divergences degrading performance on GPU, is suitable on Single Instruction Multiple Data (SIMD) execution on DSP. Among various `Optical Flow` algorithms on different platforms, only the two implementing the same Lucas Kanade algorithm are considered, where FPGA shows better performance than GPU. Notice that for `Gaussian Blur` CPU shows slightly better performance than mGPU, because multi-core is boosted for acceleration and there is no overhead for launching kernels. Table III lists preferable

TABLE III
PREFERABLE ACCELERATOR FOR CV WORKLOADS

| Workload | Gaussian Blur | Convolution | Sobel Filter | Undistort | Feature Detect | Optical Flow |
|---|---|---|---|---|---|---|
| Perf. preferable | CPU | GPU | GPU | GPU | DSP | FPGA |
| Energy preferable | mGPU | GPU | DSP | GPU | DSP | FPGA |

TABLE IV
INFERENCE LOCALLY VS. CLOUD

|  | Energy (J) | Latency (Seconds) |
|---|---|---|
| CPU | 0.800 | 0.400 |
| GPU | 0.132 | 0.033 |
| Cloud | 0.010 | 2 to 5 |

accelerator for CV workloads considering both performance and energy efficiency.

Figure 4 shows the energy consumption of CV workloads on the targeted accelerators. For CV workloads such as `Sobel Filter`, although GPU executes faster, the energy consumed is higher than DSP. FPGA also shows low energy consumption even though its execution time is longer than GPU. Overall, DSP has better energy efficiency for most CV workloads except for `Convolution`. As shown in Figure 6, GPU demonstrates the better energy efficiency in accelerating DL workloads than FPGA. While FPGA has low power consumption, the execution time is not competitive compared with GPU. To keep load balance, GPU and DSP are preferred to accelerate DL workloads and CV workloads, respectively.

**Observation Three: Setup time matters.** Figure 7 shows the breakdown of total computation offloading time of fix-point `Convolution` on accelerators including the setup time and data copying back time. As it can be seen, setup time occupies a significant portion of total time and needs to be considered in real-time applications. Even though it shows in Figure 3 that GPU has the fastest execution time for `Convolution`, when setup time is taken into account, DSP is preferable for overall execution time. GPU has the largest setup overhead even though it can be overlapped by batching, but batching is impractical in latency-critical applications. DSP has the lowest setup overhead partially because of the efficient host-to-accelerator invocation mechanism, and therefore it is suitable to accelerate lightweight workloads.

### D. Computing Locally vs. Offloading to Cloud

The energy consumption as well as the performance of running image recognition on the CPU and the GPU of the TX1 platform are compared with when offloading to the cloud using *AlexNet* [10] as the benchmark.

**Observation Four: Offloading to cloud achieves high energy efficiency but fails to meet real-time requirements.** As shown in Table IV, when executing on CPU, inferencing takes 0.4 seconds to process one image and the total energy consumed is 0.8 J; however, to perform the same task, GPU only consumes 1/6 of the energy and it is around 10 times faster compared to CPU. Offloading this task to the cloud only incurs negligible energy by only uploading the image, but the problem is that the inference latency is not only long but also highly variable, making it very hard to meet real-time requirements.

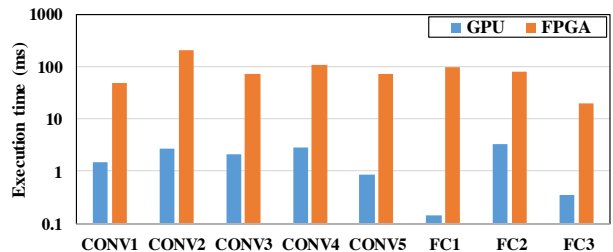

Fig. 5. DL workloads execution time: GPU outperforms FPGA for all workloads.

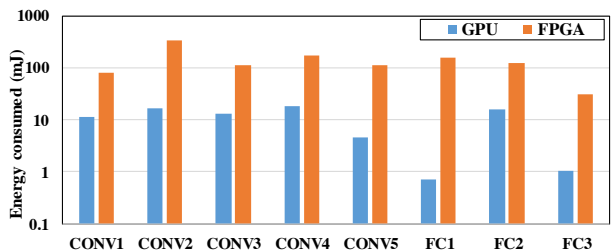

Fig. 6. DL workloads energy consumption: GPU is always more energy efficient than FPGA.

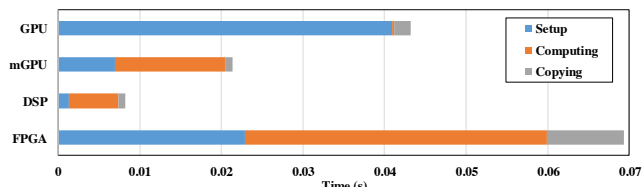

Fig. 7. Considering total computation offloading time using `Convolution`: Although GPU shows the best performance on kernel execution, the total computation time is not preferable as to DSP due to the high setup overhead.

## IV. Π-RT DESIGN

This section first lays out the design principles made from the observations in Section III, and then gives the design of the basic scheduler and the advanced scheduler in details. Through the efficient utilization of client-side heterogeneous computing resources and cloud resources, Π-RT enables full robotic workloads execution on mobile systems with stringent resource and energy constraints. Figure 8 shows the architecture of Π-RT.

With the advent of heterogeneous architectures integrating different accelerators, it is an increasing burden for the developers to manage the computing resources and to map the applications onto hardware components. To bridge the gap between robotic applications and heterogeneous hardware, Π-RT provides a transparent programming interface for users to submit tasks without knowing the details in heterogeneous hardware.

The submitted tasks are appended to the queues of different processing units (CPU, accelerators, and cloud) according to the scheduling policy. In addition, Π-RT implements a callback function for each processing unit such that when the processing unit is done with the current task, it triggers its associated callback function; then it dispatches the first task in its associated queue to the processing unit.

### A. Design Principles

The following design principles for Π-RT can be deduced from the four observations made in Section III.

- *Resource awareness*: as indicated in *Observation One*, robotic applications exhibit data-flow patterns, such that each stage depends on the output of the previous stage; execution being delayed when input is ready often leads to a waste of system resources. Since image buffers are a major consumer of system memory, the higher priority is given to tasks with image inputs in order to release the image buffers as soon as possible.
- *Heterogeneous accelerator awareness*: as indicated in *Observation Two*, no single accelerator wins all. Therefore, all accelerators available in the platform are considered instead of focusing on only one type of computing units
- *Setup time awareness*: as indicated in *Observation Three*, for each accelerator there is always a setup time associated with it. Most existing schedulers do not consider the setup overheads thus leading to non-optimal results; for Π-RT, the runtime initialization routine also initializes all the accelerators and thus minimizes the setup overhead on performance.
- *Cloud awareness*: as indicated in *Observation Four*, offloading to cloud leads to energy efficiency but fails to meet real-time requirements. The scheduler dispatches tasks without real-time requirements to cloud by default. Note that the definition of non-real-time requirement in this context is that the task does not need to complete within 30 seconds.

### B. Basic Scheduler Design

The basic scheduler does not have *resource awareness* and *cloud awareness*. To accommodate different scenarios, the basic scheduler operates in three different modes: *Latency-Optimal*, *Throughput-Optimal*, and *Energy-Optimal* as illustrated in Algorithm 1, Algorithm 2, and Algorithm 3, respectively.

**Algorithm 1** Latency-Optimal Scheduling

**Require:** A counter N for dispatched tasks; the weight of each queue: Wc, Wg, Wd.
1: **for** each waiting task **do**
2:   **if** N ¡ Wg **then** dispatch to GPU queue; N++
3:   **else if** N ¡ (Wg+Wd) **then** dispatch to DSP queue; N++
4:   **else if** N ¡ (Wg+Wd+Wc) **then** dispatch to CPU queue; N++
5:   **else** dispatch to CPU queue; reset counter N

The experimental results of executing `Convolution` with the basic scheduler deployed on Snapdragon 820 is

**Algorithm 2** Throughput-Optimal Scheduling

**Require:** The load of each queue: Qc, Qg, Qd; the weight of each queue: Wc, Wg, Wd.
1: **for** each waiting task **do**
2:   **if** Qg ¡ Wg **then** dispatch to GPU queue; Qg++
3:   **else if** Qc ¡ Wc **then** dispatch to CPU queue; Qc++
4:   **else if** Qd ¡ Wd **then** dispatch to DSP queue; Qd++
5:   **else** dispatch to CPU queue; Qc++

**Algorithm 3** Energy-Optimal Scheduling

**Require:** The load of each queue: Qc, Qg, Qd; the weight of each queue: Wc, Wg, Wd.
1: **for** each waiting task **do**
2:   **if** Qd ¡ Wd **then** dispatch to DSP queue; Qd++
3:   **else if** Qg ¡ Wg **then** dispatch to GPU queue; Qg++
4:   **else if** Qc ¡ Wc **then** dispatch to CPU queue; Qc++
5:   **else** dispatch to DSP queue; Qd++

tabulated in Table V. The scheduler checks the load of each queue and selectively appends computational tasks into different queues based on the workloads characteristics and hardware resources. Each queue size is determined according to hardware computing resources and customized to the coarse-grain profiling result of each workload. The scheduler offloads task based on Round-Robin algorithm in the *Latency-Optimal* scheduling. The *Throughput-Optimal* scheduling keeps the GPU queue full of loads to achieve high throughput. On the other hand, the *Energy-Optimal* scheduling chooses DSP over other accelerators.

Note that the queues wrap different hardware implementations of targeted workloads, the associated data transfer, and kernel launch functions. In addition, when the system starts, Π-RT performs runtime initialization to set up all accelerators, and thus hides the setup times of different accelerators.

### C. Advanced Scheduler Design

The advanced scheduler design is illustrated in Algorithm4. Π-RT implements *cloud awareness* by checking whether this task has real-time requirements; If not, the task is dispatched to the cloud. Then Π-RT implements *resource awareness* by checking whether the task consumes images as input; if so, Π-RT dispatches the task to the high priority queue. Otherwise, Π-RT falls back to the basic scheduler. Note that Π-RT always clears the high priority queue first before servicing the normal queues.

**Algorithm 4** Advanced Scheduling

**Require:** The tags of the tasks: rtt - real-time task, ii - input is an image; The load of the high priority queue: Qh.
1: **for** each waiting task **do**
2:   **if** !rtt **then** dispatch to cloud;
3:   **else if** ii **then** dispatch to the high priority queue; Qh++
4:   **else** dispatch using the basic scheduler

TABLE V
SCHEDULING ALGORITHMS COMPARISON

| Scheduling | Throughput frame per ms | Avg. latency (ms) CPU | GPU | DSP | Energy (J) |
|---|---|---|---|---|---|
| Throughput | **0.745** | 8.21 | 10.42 | - | 4.76 |
| Latency | 0.467 | **6.84** | **7.45** | 19.62 | 4.20 |
| Energy | 0.307 | - | 8.24 | 18.61 | **3.36** |

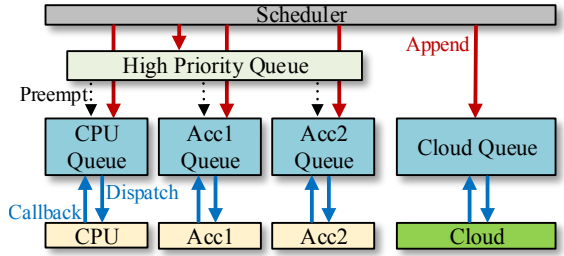

Fig. 8. The architecture of Π-RT: non-real-time tasks are appended to cloud queue; tasks with image inputs are sent to the high priority queue which will be dispatched before normal tasks.

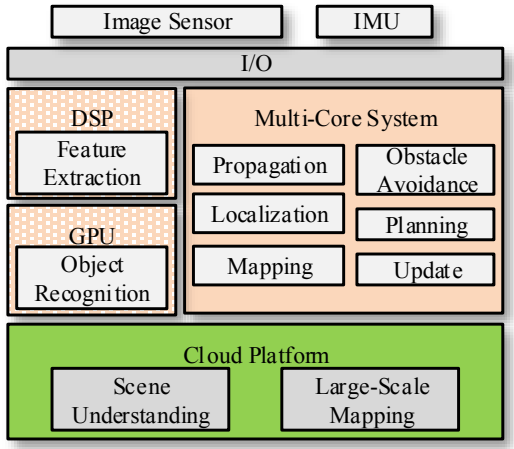

Fig. 9. Π-RT enabling full robotic workloads

## V. REAL-WORLD IMPLEMENTATION

To verify its effectiveness, Π-RT is implemented on the Snapdragon 820 SoC as shown in Table II. With Π-RT, an autonomous robot is enabled to simultaneously perform autonomous navigation, obstacle detection, route planning, large map generation, and scene understanding, all within an 11 W of computing power envelope and with negligible overhead of scheduling. Figure 9 shows a snapshot of the full robotic workloads executing on the Snapdragon 820 SoC as well as on the cloud. A video demonstration of the system can be found [11].

### A. Client-Side Performance

Without Π-RT, the robotic system (including feature extraction, localization, planning, obstacle avoidance, and deep learning) runs on CPU only. Originally, it takes 1.5 seconds to process each frame for object recognition, and the localization pipeline is only able to process 15 images per second. The feature extraction task consists of CV workloads such as `Gaussian Blur` and `Feature Detect`. Under this mode, the system stresses the CPU and keeps the four cores spinning with 10 W of power consumption on average.

After deploying Π-RT with the throughput-based scheduling scheme, the DSP is mainly utilized for sensor data processing tasks, such as feature extraction and optical flow; the GPU is mainly used for deep learning tasks, such as object recognition; the CPU cores are used to localize the vehicle, to plan paths, and to avoid obstacles in real time. The localization pipeline is able to process 25 images per second, while the deep learning pipeline is able to perform 3 object recognition tasks per second. The planning and control pipeline is designed to plan a path within 6 ms. The SoC consumes 11W of power on average and the robot is able to autonomously navigate with a max speed of 5 miles per hour.

With only 1W of extra power consumption, Π-RT achieves 60% and 3-fold performance improvement on localization and object recognition tasks by efficiently utilizing heterogeneous resources, respectively. As demonstrated through this implementation, Π-RT enables efficient utilization of heterogeneous computing units such that even mobile SoCs are capable of supporting full robotic workloads.

### B. Cloud-Side Performance

Besides the client-side execution, Π-RT is able to dispatch two non-real-time tasks to the cloud: the first is scene understanding[5] and the second one is large-scale high-precision map generation[12]. The purpose of the scene understanding is to generate a description of the scene of meta-data so that it is easier to retrieve multimedia information later; the purpose of the map generation is to perform global optimization on the visual map to improve its accuracy. Both of these tasks are not suitable for running on the client side; the scene understanding requires over 1 GB of GPU memory, and the Snapdragon 820 system simply does not have enough resource for this application; the large scale mapping involves bundle adjustment optimization over 100 images with thousands of feature points. It takes hours to run on the client side, but on the cloud side it only takes less than five minutes to finish.

## VI. RELATED WORK

Most previous works focus on either mapping of one algorithm to one type of accelerator, or on scheduling for homogeneous systems or heterogeneous systems with single accelerator. In contrast, Π-RT dynamically maps various robotic workloads to multiple accelerators and to the cloud.

**Mapping of Deep Learning and Computer Vision Workloads to Heterogeneous Architectures:** Hegde *et al.*[13] have proposed a framework for easy mapping of CNN specifications to accelerators such as FPGAs, DSPs, GPUs, and RISC multi-cores. Malik *et al.*[14] compare the performance and energy efficiency of computer vision algorithms on on-chip FPGA accelerators and GPU accelerators. Many work explored the optimization of deep learning algorithms on embedded GPU or FPGA accelerator [15], [16].

There are also many papers on optimizing computer vision related tasks on embedded platforms. Honegger *et al.*[17] propose FPGA acceleration of embedded computer vision. Satria *et al.*[18] perform platform specific optimizations of face detection on embedded GPU-based platform and reported real-time performance. Vasilyev et al.[9] evaluate computer vision algorithms on programmable architectures. Nardi et al.[19] present a benchmark suite to evaluate dense SLAM algorithms across desktop and embedded platforms in terms of accuracy, performance, and energy consumption. However, none of these work consider the complexity of integrating the various kind of workloads into a system, and only touch one component of the problem.

**Scheduling for Heterogeneous Architectures** has been broadly studied for single-ISA multiprocessors such as asymmetric multi-core architectures, i.e., big and small cores, and multi-ISA multiprocessors such as CPU with GPU.

On the single-ISA multiprocessors side, many work have been done at the operating system level to map workload onto most appropriate core type in run time. Koufaty *et al.*[20] identified that the periods of core stalls is a good indicator to predict the core type best suited for an application. Based on the indicator, it added a biased schedule strategy on operating systems to improve system throughput. Saez *et al.*[21] proposed a scheduler that add efficiency specialization and TLP (thread-level parallelism) specialization in operating systems to optimize throughput and power at the same time. Efficient specialization maps CPU-intensive workloads onto fast cores and memory-intensive workloads onto slow cores. TLP specialization uses fast cores to accelerate sequential phase of parallel applications and use slow cores for parallel phase to achieve energy efficiency.

On the asymmetric multi-core architectures side, Jimenez *et al.*[22] proposed a user-level scheduler for CPU with GPU like system. It evaluates and records the performance of a process on each PE at the initial phase. Then, based on this history information, it maps the application onto the best suited PE. Luk *et al.*[23] focus on improving the latency and energy consumption of a single process. It uses dynamic compilation to characterize workloads, determines optimal mapping and generates codes for CPUs and GPUs.

**Π-RT is the first runtime framework that is designed for efficient execution of robotic workloads on both client-side heterogeneous architectures as well as the cloud; it has been implemented in production robotic systems**. The design principles, including *resource-awareness*, *heterogeneous-awareness*, *setup-time-awareness*, and *cloud awareness* are deduced from practical experiences and have proven to achieve better performance and energy efficiency.

## VII. CONCLUSIONS

Efficient utilization of client-side heterogeneous computing resources as well as the cloud is the key to enable full robotic workloads on embedded systems. In this paper, we first conduct a comprehensive study of emerging robotic applications on heterogeneous SoC architectures. Based on the results, we design and implement Π-RT, the first robotic runtime framework which efficiently utilizes not only the on-chip heterogeneous computing resources but also the cloud to achieve high performance and energy efficiency. Π-RT is deployed on a production mobile robot to demonstrate that full robotic workloads, including autonomous navigation, obstacle detection, route planning, large map generation, and scene understanding, can be efficiently executed simultaneously with an 11 W of power consumption.